\pdfoutput=1

\documentclass[11pt]{article}

\usepackage{emnlp2021}

\usepackage{times}
\usepackage{latexsym}

\usepackage[T1]{fontenc}

\usepackage[utf8]{inputenc}

\usepackage{microtype}
\usepackage{graphicx}
\usepackage{amsmath}
\title{Speaker Turn Modeling for Dialogue Act Classification}

\author{
  Zihao He$^{1,3}$ \quad Leili Tavabi$^{1,2}$ \quad Kristina Lerman$^{3}$ \quad Mohammad Soleymani$^{2}$ \\
  $^1$Department of Computer Science, University of Southern California \\
  $^2$Institute for Creative Technologies, University of Southern California \\
  $^3$Information Sciences Institute, University of Southern California\\
  \footnotesize \texttt{ zihaoh@usc.edu} \quad \texttt{ltavabi@ict.usc.edu}  \quad \texttt{lerman@isi.edu}  \quad  \texttt{soleymani@ict.usc.edu} \\
  }

\begin{document}
\maketitle

\begin{abstract}
Dialogue Act (DA) classification is the task of classifying utterances with respect to the function they serve in a dialogue. Existing approaches to DA classification model utterances without incorporating the turn changes among speakers throughout the dialogue, therefore treating it no different than non-interactive written text. In this paper, we propose to integrate the turn changes in conversations among speakers when modeling DAs. Specifically, we learn conversation-invariant speaker turn embeddings to represent the speaker turns in a conversation; the learned speaker turn embeddings are then merged with the utterance embeddings for the downstream task of DA classification. With this simple yet effective mechanism, our model is able to capture the semantics from the dialogue content while accounting for different speaker turns in a conversation. Validation on three benchmark public datasets demonstrates superior performance of our model.\footnote{Our code and data are publicly available at \url{https://github.com/ZagHe568/speak-turn-emb-dialog-act-clf}.}
\end{abstract}

\section{Introduction}
\label{sec:intro}
Dialogue Acts (DAs) are the functions of utterances in the context of a dialogue conveying the speaker's intent \cite{searle1969speech}. 
In natural language understanding, DA classification is of critical importance, as it underlies various tasks such as dialogue generation \cite{li2017adversarial} and intention recognition \cite{higashinaka2006incorporating}, 
thus providing effective means for domains like dialogue systems \cite{higashinaka2014towards}, talking avatars \cite{xie2014statistical} and therapy \cite{xiao2016behavioral,tavabi2021analysis,tavabi2020multimodal}.


Recent studies of DA classification have leveraged deep learning techniques, where promising results have been observed. Generally, these methods utilize hierarchical Recurrent Neural Networks (RNNs) to model structural information between utterances, words, and characters \cite{raheja2019dialogue, li2018dual, wan2018improved, chen2018dialogue, kumar2018dialogue, bothe2018context}. However, most of these approaches treat a spoken dialogue similar to written text, thereby neglecting 
to explicitly model turn-taking across different speakers.
Inherently, computational understanding of a dialogue, which has been generated by multiple parties with different goals and habits in an interactive and uncontrolled environment \cite{chi2017speaker}, requires modeling turn-taking behavior and temporal dynamics of a conversation. For instance, in a dyadic conversation, given an utterance with dialogue act ``Question'' from speaker A, if the following utterance is from speaker B, then the corresponding act is likely to be ``Answer''; however, if there is no change in speakers, then the following act is less likely to be ``Answer.'' Therefore, modeling turn changes in conversations is essential.

In this regard, we aim to incorporate the speaker turns into encoding an utterance.
Specifically, we propose to model speaker turns in conversations and introduce two speaker turn embeddings that are combined with the utterance embeddings.
Given a conversation containing a sequence of utterances, we first obtain the utterance embeddings using a large pretrained language model RoBERTa \cite{liu2019roberta}, and extracting the [CLS] token embeddings from the last layer; meanwhile we use a speaker turn embedding layer to generate speaker turn embeddings given the speaker labels. The speaker turn embeddings are added to the utterance embeddings to obtain speaker turn-aware utterance representations. These representations are fed into an RNN to encode the context of the conversation, where the output hidden states are used for DA classification. We evaluate the proposed method on three benchmark datasets and achieve the state-of-the-art results, among all inductive learning methods, on two of the datasets. We argue that this simple technique provides effective means for obtaining more powerful representations for dialogue.

\section{Related Work}

\textbf{Dialogue Act Classification.} 
\citet{chen2018dialogue} propose a CRF-attentive structured network and apply structured attention network to the CRF (Conditional Random Field) layer in order to simultaneously model contextual utterances and the corresponding DAs. 
\citet{li2018dual} introduce a dual-attention hierarchical RNN to capture information about both DAs and topics, where the best results are achieved by a transductive learning model. 
\citet{raheja2019dialogue} utilize a context-aware self-attention mechanism coupled with a hierarchical RNN. 
\citet{colombo2020guiding} leverage the seq2seq model to learn the global tag dependencies instead of the widely used CRF that captures local dependencies; this method, however, requires beam search that introduces more complexity.
The aforementioned methods are based on hierarchical RNNs and neglect speaker turns modelled in this paper.

\textbf{Speaker Role Modeling in Dialogues.} 
Existing work mainly focus on speaker roles for the purpose of encoding dialogue context in conversations, involving distinguishable speaker roles like guide versus tourist.
For encoding role-based context information, \citet{chi2017speaker} and \citet{chen2017dynamic} use individual recurrent modules for each speaker role, modeling the role-dependent goals and speaking styles, and taking the sum of the resulting representations from each speaker. Similarly, \citet{hazarika2018conversational} obtain history context representations per speaker by modeling separate memory cells using Gated Recurrent Units (GRUs) for each speaker; therefore speaker-based histories undergo identical but separate computations before being combined for the downstream task.
\citet{qin2021co} treat an utterance as a vertex and add an edge between utterances of the same speakers to construct cross-utterances connections; such connections are based on specific speaker roles.
Different from speaker role-based methods, our method focuses on speaker turns and thus is still useful when speakers are not associated with specific roles. Additionally, previous methods incorporate speaker information by proposing more complex and specialized models, which inevitably introduce a large number of parameters to train, whereas we introduce two global \emph{additive} embedding vectors, requiring negligible modifications to a recurrent model and introducing $O(1)$ space complexity, as can be seen in Section \ref{sec:speaker-turn}.

\section{Methods}
\begin{figure}[ht]
    \centering
    \includegraphics[width=0.45\textwidth]{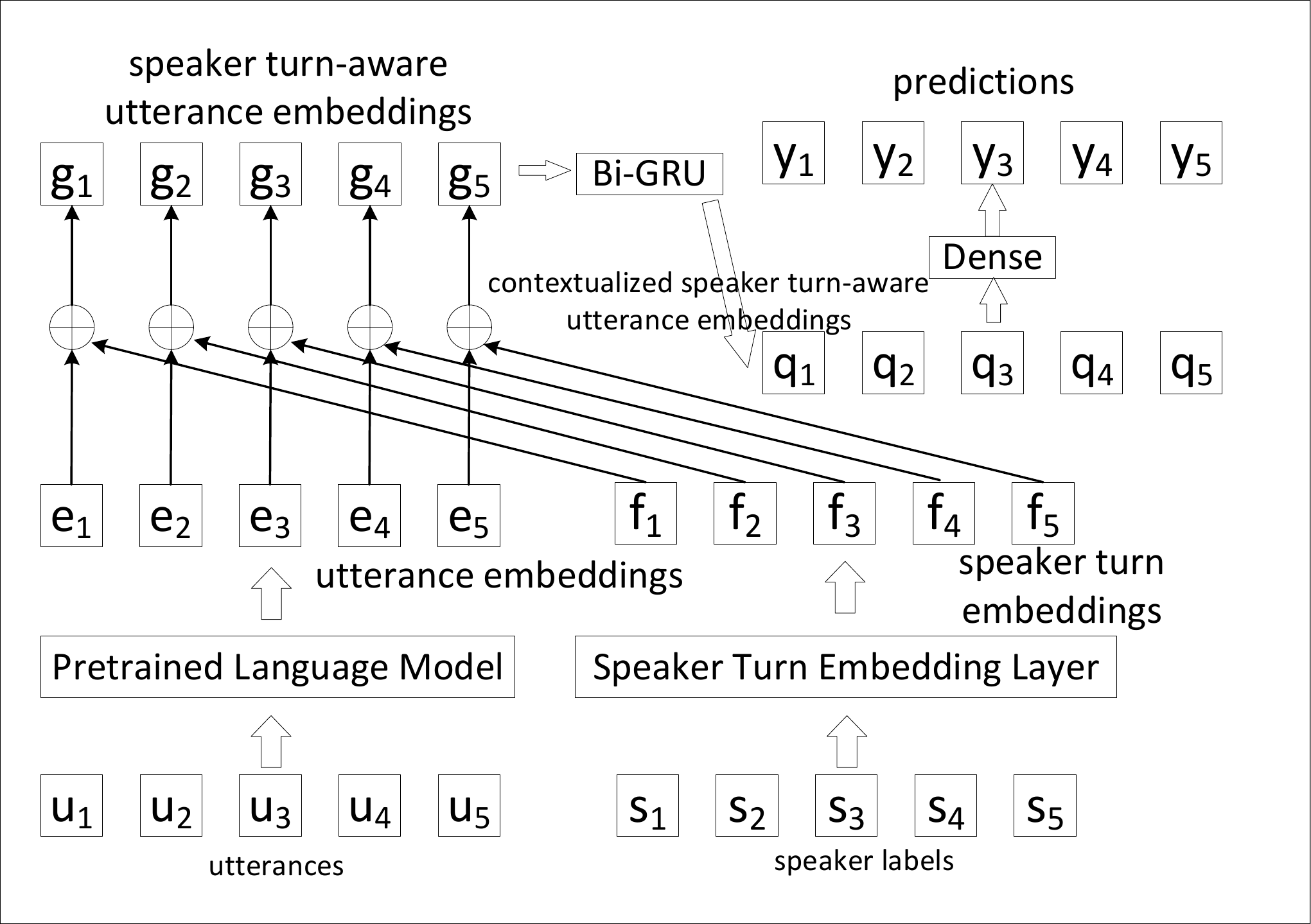}
    \caption{The overall framework of our proposed method. In this toy example, the conversation consists of five utterances.}
    \label{fig:archi}
\end{figure}

The overall framework of our model is shown in Figure \ref{fig:archi}.  In this section, we will describe our model's components in detail.

\subsection{Problem Definition}
The input corpus $D=\{(C_n, Y_n, S_n)\}_{n=1}^N$ consists of $N$ conversations, where $C_n=\langle u_t^n\rangle_{t=1}^T$ is a dialogue instance containing a sequence of $T$ utterances, $Y_n=\langle y_t^n\rangle_{t=1}^T$ and $S_n=\langle s_t^n\rangle_{t=1}^T$ are the corresponding DA labels and speaker labels. The goal is to learn a model from corpus $D$, such that given an unseen conversation $C_p$ and its corresponding speaker labels $S_p$, the model is able to predict the DA labels $Y_p$ of utterances in $C_p$.

\subsection{Utterance Modeling}
We use the pretrained language model RoBERTa to encode utterances, which enables us to utilize the powerful representations obtained from pretrainining on large amounts of data. Given an utterance $u$, we take the embedding of [CLS] token from the last layer as the utterance embedding, denoted as $e(u)$.

\subsection{Speaker Turn Modeling}
\label{sec:speaker-turn}
Different from text written by a single author in a non-interactive environment, dialogues usually involve multiple parties, in minimally-controlled environments where each speaker has their own goals and speaking styles \cite{chi2017speaker}. Therefore, it is critically important to model how speakers take turns individually and inform the model when there is a speaker turn change. To this end, for a dyadic conversation corpus, we introduce two conversation-invariant speaker turn embeddings, for each interlocutor. These two embeddings are trained across all speakers in the train set and are independent of any given conversation or speaker pair. The two embeddings are learnable parameters during the optimization and have the same size as the utterance embeddings, which are generated by a speaker turn embedding layer with speaker labels as input. Note that in a dyadic conversation, speaker labels ($0/1$) naturally indicate speaker turn changes.
This idea is inspired by the positional encoding in Transformers \cite{vaswani2017attention}, where the authors introduce a positional embedding with the same size as the token embedding at each position, and the positional embeddings are shared across different input sequences. 
For a multi-party conversation corpus, because our goal is to model speaker turns instead of assigning a different embedding to each speaker, we relabel the speakers and flip the speaker label (from 0 to 1 and vice versa) when there is speaker turn change; for example, if the original speaker sequence is $\langle 0,0,1,2,3,3,1\rangle$, after relabeling it becomes $\langle0,0,1,0,1,1,0\rangle$, which can then be represented by the two introduced speaker turn embeddings. This simplifies turn-change modeling, as the number of speakers in different conversations can be different.

Encoding speaker turns instead of individual speaker styles/characteristics provides the following advantages: 1) in datasets with many different speakers across relatively short dialogue sessions, it is challenging to transfer the learned speaker representations across different sessions; 2) the simplicity of this mechanism makes it more scalable for multi-party dialogue sessions with larger number of speakers.

To obtain the speaker turn-aware utterance embedding $g(u,s)$, given an utterance $u$ and its binary speaker turn label $s$, the speaker turn embedding $f(s)$ is then added to the utterance embedding $e(u)$, such that
    $g(u,s) = e(u) + f(s), s \in \{0, 1\}$.
The idea of taking the sum is also inspired by Transformers where they add the positional embeddings to token embeddings for sequence representation \cite{vaswani2017attention}.
We also considered the concatenation of the speaker turn embedding and the utterance embedding, resulting in inferior performance compared to taking the sum.

\subsection{Conversational Context Modeling}
Context plays an important role in modeling dialogue, which should be taken into account when performing DA classification. Given a sequence of independently encoded speaker turn-aware utterance embeddings $\langle g(u_t, s_t)\rangle_{t=1}^n$ in conversation $C$, we used a Bi-GRU \cite{cho2014properties} to inform each utterance of its context, such that
    $\langle q(u_t, s_t)\rangle_{t=1}^n = \text{GRU}\langle g(u_t, s_t)\rangle_{t=1}^n$,
where $\langle q(u_t, s_t)\rangle_{t=1}^n$ are contextualized speaker turn-aware utterance embeddings from the hidden states of the Bi-GRU model. These embeddings are then fed into a fully connected layer for DA classification, which is optimized using a cross-entropy loss. Different from existing work \cite{raheja2019dialogue, li2018dual, wan2018improved, chen2018dialogue, kumar2018dialogue, bothe2018context}, we do not use a CRF layer in our method, because our experiments indicate that it brings modest performance gains at the expense of adding more complexity.

\section{Experiments and Results}
\subsection{Datasets}
We evaluate the performance of our model on three public datasets: the Switchboard Dialogue Act Corpus (SwDA) \cite{jurafsky1997switchboard, shriberg1998can, stolcke2000dialogue}, the Meeting Recorder Dialogue Act Corpus (MRDA) \cite{shriberg2004icsi}, and the Dailydialog (DyDA) \cite{li2017dailydialog}. \textbf{SwDA}\footnote{https://github.com/cgpotts/swda} contains dyadic telephone conversations labeled with 43 DA classes; the conversations are assigned to 66 manually-defined topics. \textbf{MRDA}\footnote{https://github.com/NathanDuran/MRDA-Corpus} consists of multi-party meeting conversations and 5 DA classes. \textbf{DyDA}\footnote{http://yanran.li/dailydialog} corpus consists of human-written daily dyadic conversations labeled with 4 DA classes; the conversations are assigned to 10 topics. For SwDA and MRDA, we use the train, validation and test splits following \cite{lee2016sequential}. For DyDA, we use its original splits \cite{li2017dailydialog}. The statistics of the three datasets are summarized in Table \ref{tab:datasets}.

\begin{table}[ht]
\centering
\begin{small}
\addtolength{\tabcolsep}{-3pt}
\begin{tabular}{cccccc}
\hline
Dataset & $|C|$ & $|P|$   & Train     & Val      & Test    \\ \hline
SwDA    & 43  & 2     & 1003/193K & 112/20K  & 19/4.5K \\
MRDA    & 5   & multiple & 51/75k    & 11/15.3K & 11/15K  \\
DyDA    & 5   & 2     & 11K/87.1K & 1K/8K    & 1K/7.7K \\ \hline
\end{tabular}
\addtolength{\tabcolsep}{3pt}
\end{small}
\caption{The statistics of the three datasets. $|C|$ denotes the number of DA classes; $|P|$ denotes the number of parties; Train/Val/Test denotes the number of conversations/utterances in the corresponding split.}
\label{tab:datasets}
\end{table}

\subsection{Experimental Setup}
\begin{figure}[ht]
    \centering
    \includegraphics[width=0.48\textwidth]{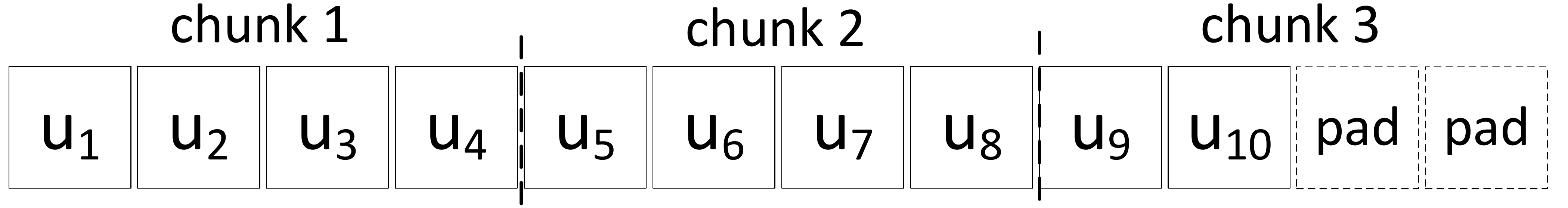}
    \caption{A toy example of slicing a conversation of length 10 into 3 chunks of length of 4.}
    \label{fig:chunks}
\end{figure}

On SwDA and DyDA, which are two dyadic conversation corpora, we use the original speaker labels (due to equivalence to speaker turn change labels); however, since MRDA is a multi-party conversation corpus, we use the binary speaker turn change labels obtained from the sequence of speaker labels as mentioned in Section \ref{sec:speaker-turn}. 
On DyDA, because the maximum length of conversations (number of utterances) is less than 50, we treat each conversation as a data point and pad all conversations to the maximum length. However, conversations in SwDA and MRDA are much lengthier (up to 500 in SwDA and 5,000 in MRDA); to avoid memory overflow when training on a GPU, we slice the conversations into shorter fixed-length chunk sizes of 128 and 350 for SwDA and MRDA respectively, as shown in Figure \ref{fig:chunks}, where each chunk would represent a data point. 
The slicing operation is only needed for training but not in the validation or test, because in training a computation graph is maintained which consumes significantly more GPU memory.
More details about the setup is reported in Appendix A. The results using different chunk sizes are reported in Appendix B.

\subsection{Baselines}
We consider deep learning based approaches as baselines including DRLM-Cond \cite{ji2016latent}, Bi-LSTM-CRF \cite{kumar2018dialogue}, CRF-ASN \cite{chen2018dialogue}, ALDMN \cite{wan2018improved}, SelfAtt-CRF \cite{raheja2019dialogue}, SGNN \cite{ravi2018self}, DAH-CRF-Manual \cite{li2018dual}, and Seq2Seq \cite{colombo2020guiding}. We report the results of DRLM-Cond and Bi-LSTM-CRF on DyDA implemented by \cite{li2018dual}. Our proposed speaker turn modeling is usable in other embedding-based approaches to DA classification, but because none of the recently published work have made the code available, we do not implement the proposed speaker turn modeling on top of the baselines.

For fair comparison with DAH-CRF-Manual$_{conv}$ \cite{li2018dual} where manual conversation-level topic labels are used, we assign all utterances in a conversation the corresponding conversation topic label.
To utilize the topic information, following the idea of speaker turn embedding in Section \ref{sec:speaker-turn}, we introduce an embedding $h(m)$ for each topic $m$ and add it to the speaker turn-aware utterance embedding, such that
$l(u,s,m) = g(u,s) + h(m)$
where $l(u,s,m)$ is the obtained speaker turn and topic-aware utterance embedding.

Note that we do not compare our results to DAH-CRF-LDA$_{conv}$ and DAH-CRF-LDA$_{utt}$ \cite{li2018dual}, which are categorized as transductive learning because they utilize the data from training, validation and test sets to perform LDA topic modeling and use the learned topic labels to supervise the training process. In contrast, our method and all baselines are categorized as inductive learning, which do not use supervision from the validation or test set. In addition, we do not compare to Seq2Seq \cite{colombo2020guiding} on SwDA where they adopt a different test split from the one used in our method and the baselines.

\subsection{Results}
The results from our method and the baselines are shown in Table \ref{tab:results}. Our method achieves state-of-the-art results on SwDA and DyDA; on MRDA it achives the performance comparable to the state-of-the-art.
Notably, on SwDA and MRDA, comparing the proposed model (Ours) to the model without speaker turn embeddings (Ours$\neg$Speaker),
we observe significant improvements in performance, signifying the effectiveness of modeling speaker turns in dialogue representation. 
On DyDA, the performance gains slightly after applying speaker turn modeling; we argue that this is because in conversations in DyDA, there is a consistent speaker turn change after each utterance following the pattern $\langle 0,1,0,1,0, 1\rangle$; such a pattern is more predictable, and therefore, modeling speaker turns provides limited auxiliary information, from the perspective of information theory.

\begin{table}[ht]
\centering
\begin{small}
\begin{tabular}{lccc}
\hline
\textbf{Dataset} & \textbf{SwDA} & \textbf{MRDA} & \textbf{DyDA} \\ \hline
DRLM-Cond       & 77.0          & 88.4          & 81.1          \\
Bi-LSTM-CRF     & 79.2          & 90.9          & 83.6          \\
CRF-ASN         & 80.8          & 91.4 & -             \\
ALDMN           & 81.5          & -             & -             \\
SelfAtt-CRF     & 82.9          & 91.1          & -             \\
SGNN            & 83.1          & 86.7          & -             \\
DAH-CRF-Manual   & 80.9          & -             & 86.5          \\ 
Seq2Seq            & -          & \textbf{91.6}          & -             \\ \hline
Ours$\neg$Speaker & 82.4          & 90.7          & 86.8          \\
Ours             & \textbf{83.2} & 91.4 & 86.9          \\
Ours+Topic      & 82.4             & -             & \textbf{87.5} \\ \hline
\end{tabular}
\end{small}
\caption{Results of DA classification on three different methods. ``Ours$\neg$Speaker'' represents our method without adding speaker turn embeddings; ``Ours+Topic'' represents the proposed method using speaker turn and topic-aware embeddings for fair comparison to baselines utilizing topic information. State-of-the-art results are highlighted in bold.}
\label{tab:results}
\end{table}

In addition, on DyDA, the model Ours$\neg$Speaker outperforms the baselines, although this is not observed on SwDA and MRDA. We hypothesize that the reason may be from the fact that RoBERTa \cite{liu2019roberta} is pretrained on a large corpus of written text, which will make it better for processing the human-written conversations in DyDA, in comparison to the transcripts of telephone conversations and meeting records in SwDA and MRDA. As a result, the generated utterance embeddings are of higher quality, leading to the high performance of Ours$\neg$Speaker on DyDA.

In terms of modeling topics, on DyDA, topic information significantly improves the classification performance; in contrast, on SwDA, the performance suffers when utilizing topic information, as can be observed from the comparison of Ours and Ours+Topic.
Therefore, leveraging topic labels does not consistently lead to performance improvement; on the other hand, it is consistently improved by encoding speaker turn changes on all three datasets.


\section{Conclusion and Future Work}
In this paper, we propose a model for encoding speaker turn changes to tackle DA classification. Specifically, we introduce conversation-invariant speaker turn embeddings and add them to utterance embeddings produced by a pretrained language model. Such a simple yet scalable module can be easily added to other models to obtain significantly better results. Experiments on three datasets demonstrate the effectiveness of our method. For future work, we will explore transformer encoders \cite{vaswani2017attention} instead of RNNs for encoding context, since they have shown to be advantageous in performance and training time. 
Our improved representations can be further utilized in other downstream tasks involving dialogue, including speaker intent classification. Our findings motivate future work on encoding other interactive aspects of dialogue data into existing text representations. 
\section*{Acknowledgements}
Research was sponsored in-part by the Army Research Office and was accomplished under Cooperative Agreement Number W911NF-20-2-0053. The views and conclusions contained in this document are those of the authors and should not be interpreted as representing the official policies, either expressed or implied, of the Army Research Office or the U.S. Government. The U.S. Government is authorized to reproduce and distribute reprints for Government purposes notwithstanding any copyright notation herein.
\bibliography{anthology,custom}
\bibliographystyle{acl_natbib}

\clearpage
\appendix


\section{Experimental Setup}
The maximum feasible chunk sizes without a CUDA memory overflow, on our machines with 11GB of RAM, are 300(>128) and 700(>350) on SwDA and MRDA respectively, which indicates that using the entire un-sliced conversation is not necessary and will lead to performance deterioration due to the gradient vanishing and gradient explosion problems in RNN. 

We implement our model using PyTorch and train our model using Adam optimizer on 2 GTX 1080Ti GPUs. On SwDA and MRDA, we use a batch size of 2; and on DyDA, the batch size is 10. All batch sizes are the maximum before a memory overflow happens.
On all three datasets, we use a learning rate of $1e-4$ and train the model for a maxium of 50 epochs and report the test accuracy in the epoch where the best validation accuracy is achieved. The running time for an epoch are \textasciitilde20min, \textasciitilde5min, and \textasciitilde45min on SwDA, MRDA and DyDA respectively.

\section{Effect of Chunk Sizes}
Keeping other hyperparameters unchanged, we show the results of using different chunk sizes on SwDA and MRDA in Table \ref{tab:chunk-swda} and Table \ref{tab:chunk-mrda} respectively. On both datasets, with the chunk size increasing from a small value, the performance increases, where more context information is available to the RNN to leverage. However, after a certain value, the performance deteriorates as the chunk size further increases, in which case the gradient vanishing and gradient explosion happens in RNN and it forgets the long-term dependencies. Therefore, we argue that in order to achieve better performance in DA classification, taking the holistic conversation as input leads to inferior performance compared to slicing a long conversation into shorter chunks.

\begin{table}[ht]
\centering
\begin{tabular}{ccccc}
\hline
chunk\_size & 32   & 64   & 85   & 128  \\
acc        & 82.9 & 82.7 & 82.8 & 83.2 \\ \hline
chunk\_size & 160  & 196  & 256  & 300  \\
acc        & 82.7 & 83.0 & 82.9 & 82.3 \\ \hline
\end{tabular}
\caption{The accuracies using different chunk sizes on SwDA.}
\label{tab:chunk-swda}
\end{table}

\begin{table}[ht]
\centering
\begin{tabular}{ccccc}
\hline
chunk\_size & 85   & 175  & 350  & 700  \\
acc         & 91.3 & 91.1 & 91.4 & 91.3 \\ \hline
\end{tabular}
\caption{The accuracies using different chunk sizes on MRDA.}
\label{tab:chunk-mrda}
\end{table}

\end{document}